\PassOptionsToPackage{prologue,dvipsnames}{xcolor}
\documentclass[sigconf]{acmart}

\AtBeginDocument{%
  \providecommand\BibTeX{{%
    \normalfont B\kern-0.5em{\scshape i\kern-0.25em b}\kern-0.8em\TeX}}}

\copyrightyear{2023} 
\acmYear{2023} 
\setcopyright{acmcopyright}\acmConference[MM '23]{Proceedings of the 31st ACM International Conference on Multimedia}{October 29--November 2, 2023}{Ottawa, Canada}
\acmBooktitle{Proceedings of the 30th ACM International Conference on Multimedia
(MM '23), October 29--November 2, 2023, Ottawa, Canada}
\acmPrice{15.00}

\acmSubmissionID{80}

\usepackage{multirow}
\usepackage{algorithm}
\usepackage{algorithmicx}
\usepackage{algpseudocode}
\usepackage{amsmath}
\usepackage{url}
\usepackage{soul}
\usepackage{balance}

\usepackage{amssymb}
\usepackage{booktabs}
\usepackage[percent]{overpic}
\usepackage{makecell}

\usepackage{enumitem}
\usepackage[dvipsnames]{xcolor}

\usepackage[capitalize]{cleveref}
\crefname{section}{Sec.}{Secs.}
\Crefname{section}{Section}{Sections}
\Crefname{table}{Table}{Tables}
\crefname{table}{Tab.}{Tabs.}

\copyrightyear{2023} 
\acmYear{2023} 
\setcopyright{acmlicensed}\acmConference[MM '23]{Proceedings of the 31st ACM International Conference on Multimedia}{October 29-November 3, 2023}{Ottawa, ON, Canada}
\acmBooktitle{Proceedings of the 31st ACM International Conference on Multimedia (MM '23), October 29-November 3, 2023, Ottawa, ON, Canada}
\acmPrice{15.00}
\acmDOI{10.1145/3581783.3611707}
\acmISBN{979-8-4007-0108-5/23/10}

\begin{document}

\title[Neural Explicit Surface]{Explicifying Neural Implicit Fields for Efficient Dynamic Human Avatar Modeling via a Neural Explicit Surface}



\author{Ruiqi Zhang}
\orcid{0000-0003-0095-3672}
\affiliation{%
  \institution{Department of Computer Science, Hong Kong Baptist University}
  \streetaddress{Kowloon Tong}
  \country{Hong Kong SAR, China}
}
\email{csrqzhang@comp.hkbu.edu.hk}

\author{Jie Chen}
\orcid{0000-0001-8419-4620}
\authornote{Corresponding author: Jie Chen}
\affiliation{%
  \institution{Department of Computer Science, Hong Kong Baptist University}
  \streetaddress{Kowloon Tong}
  \country{Hong Kong SAR, China}
}
\email{chenjie@comp.hkbu.edu.hk}

\author{Qiang Wang}
\orcid{0000-0002-2986-967X}
\affiliation{%
  \institution{School of Computer Science and Technology, Harbin Institute of Technology, Shenzhen}
  \streetaddress{Kowloon Tong}
  \country{Shenzhen, China}
}
\email{qiang.wang@hit.edu.cn}

\renewcommand{\shortauthors}{Zhang et al.}

\begin{abstract}
This paper proposes a technique for efficiently modeling dynamic humans by explicifying the implicit neural fields via a Neural Explicit Surface (NES). Implicit neural fields have advantages over traditional explicit representations in modeling dynamic 3D content from sparse observations and effectively representing complex geometries and appearances. Implicit neural fields defined in 3D space, however, are expensive to render due to the need for dense sampling during volumetric rendering. Moreover, their memory efficiency can be further optimized when modeling sparse 3D space. To overcome these issues, the paper proposes utilizing Neural Explicit Surface (NES) to explicitly represent implicit neural fields, facilitating memory and computational efficiency. To achieve this, the paper creates a fully differentiable conversion between the implicit neural fields and the explicit rendering interface of NES, leveraging the strengths of both implicit and explicit approaches. This conversion enables effective training of the hybrid representation using implicit methods and efficient rendering by integrating the explicit rendering interface with a newly proposed rasterization-based neural renderer that only incurs a texture color query once for the initial ray interaction with the explicit surface, resulting in improved inference efficiency. NES describes dynamic human geometries with pose-dependent neural implicit surface deformation fields and their dynamic neural textures both in 2D space, which is a more memory-efficient alternative to traditional 3D methods, reducing redundancy and computational load. The comprehensive experiments show that NES performs similarly to previous 3D approaches, with greatly improved rendering speed and reduced memory cost.
\end{abstract}

\ccsdesc[500]{Computing methodologies~Volumetric models}
\ccsdesc[300]{Computing methodologies~Animation}
\ccsdesc[300]{Computing methodologies~Mesh models}

\keywords{Dynamic Human Modelling, Neural Fields}

\maketitle

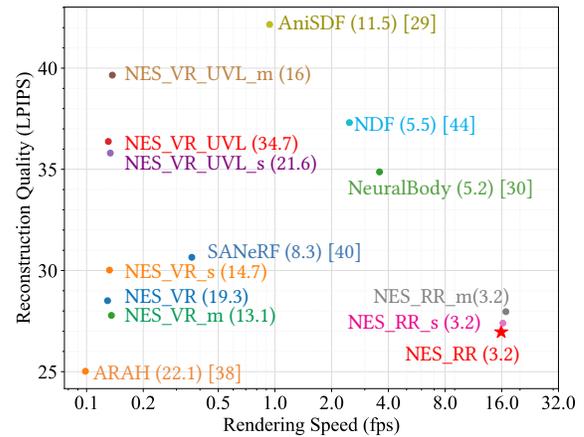
\begin{figure}[t!]
\begin{overpic}[width=1.0\linewidth]{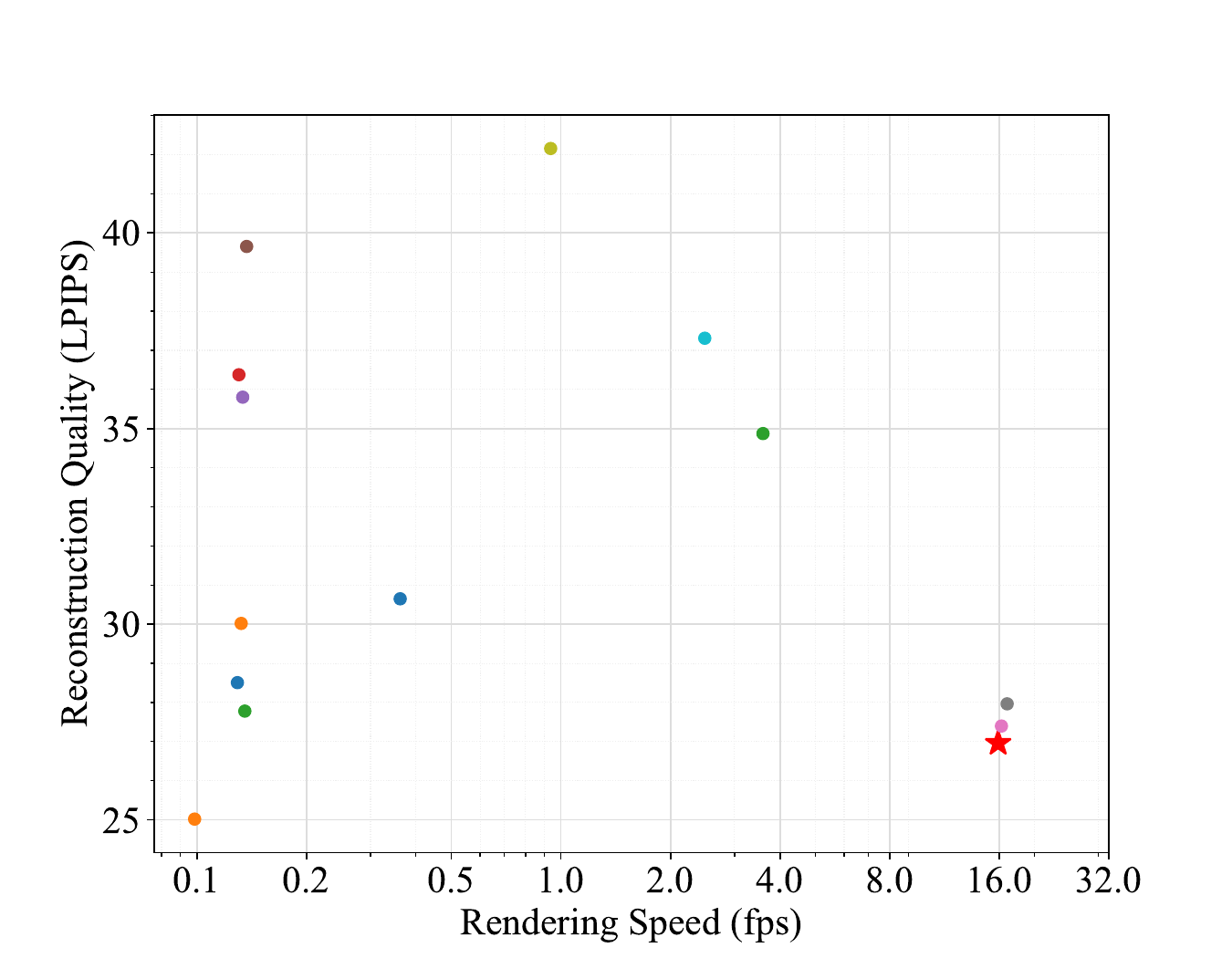}

\put(61,22){\color{gray}\small NES\_RR\_m(3.2)}
\put(57,18){\color{magenta}\small NES\_RR\_s (3.2)}
\put(66,13){\color{red} \small NES\_RR (3.2)}

\put(22,46){\color{red}\small NES\_VR\_UVL (34.7)}
\put(22,43){\color{violet}\small NES\_VR\_UVL\_s (21.6)}
\put(22,57){\color{brown}\small NES\_VR\_UVL\_m (16)}

\put(22,26){\color{orange}\small NES\_VR\_s (14.7)}
\put(22,22){\color{NavyBlue}\small NES\_VR (19.3)}
\put(22,19){\color{ForestGreen}\small NES\_VR\_m (13.1)}

\put(17,10){\textcolor[RGB]{239,134,54}{\small ARAH~(22.1)~\cite{wang2022arah}}}
\put(35,29){\textcolor[RGB]{59,117,175}{\small SANeRF~(8.3)~\cite{xu2022surface}}}
\put(57,39){\textcolor[RGB]{81,158,62}{\small NeuralBody~(5.2)~\cite{peng2021neural}}}

\put(46,65){\color{olive}\small AniSDF~(11.5)~\cite{peng2022animatable}}
\put(58,49){\color{cyan}\small NDF~(5.5)~\cite{zhang2022ndf} }
\end{overpic}
\vspace{-0.5cm}
\caption{Rendering Efficiency. The values in brackets represent runtime memory cost (in Gb). The horizontal axis shows rendered Frames per Second (FPS) on a log scale. The vertical axis represents LPIPS metrics. Our method, NES\_RR, achieves the best balance between rendering efficiency, reconstruction quality, and memory cost. Though ARAH yields slightly better LPIPS than our NES\_RR, its memory cost is seven times greater than ours and its render speed is over 160 times slower than our NES\_RR.}
\vspace{-0.5cm}
\label{fig:efficiency}
\end{figure}
\section{Introduction}
\label{sec:intro}

Dynamic human modeling has attracted extensive research interest due to its wide range of potential applications. It has been employed to facilitate film productions, such as the iconic bullet time effect in The Matrix, and to realize telepresence, which significantly improves communication efficiency. In addition, the upcoming metaverse technology could leverage dynamic human modeling to create a person-specific avatar that brings more fidelity to the virtual world.

Despite its wide range of applications, modeling dynamic humans has remained a challenging problem for decades. In the traditional approach, which involves reconstructing geometry and texture, the process requires dense camera rigs~\cite{gortler1996lumigraph,hedman2018deep,joo2018total} or controlled lighting conditions~\cite{collet2015high,guo2019relightables} to capture the human model. Skilled artists then manually create a skeleton for the human model and meticulously design skinning weights to achieve realistic animation~\cite{lewis2000pose}. However, these systems are often bulky and expensive, and require professional operators. As a result, such models are limited to simple skinning blending techniques, restricting them from achieving more complex or non-linear deformations.

Neural fields have recently emerged as a powerful tool for high-quality modeling of dynamic humans, enabling the representation of objects in 3D space with neural networks recording the scene status at each 3D location. NeRF~\cite{mildenhall2020nerf}, for instance, models static scenes using MLP networks and captures the scene status for each 3D location in terms of density and color. The densities and colors of sample points along a ray are then aggregated using a differentiable volumetric renderer, which can be directly learned from multi-view images. To build on the success of directly learned neural fields from multi-view images, many researchers have proposed novel frameworks to reduce costs and improve flexibility in modeling dynamic humans. While some approaches~\cite{saito2019pifu, saito2020pifuhd, he2020geo} reconstruct humans directly from images, they lack control over human poses and generalization to out-of-distribution poses. On the other hand, other methods~\cite{peng2021animatable, weng2022humannerf, zhang2022ndf, xu2022surface} aggregate information across video frames and demonstrate better ability to handle novel poses.

Among these methods, our approach focuses on reconstruction from video and models the dynamic human as pose-conditioned neural fields~\cite{zhang2022ndf, xu2022surface}. This technique enables texture colors to change according to pose changes, improving fidelity. While other methods~\cite{peng2021animatable, weng2022humannerf, geng2023learning, li2022tava} establish dense correspondences between video frames through deformation fields for joint learning of a shared canonical model, they can neglect variations in lighting or wrinkle shadows, which can decrease fidelity. In contrast, our approach leverages pose-conditioned neural fields to achieve efficient and high-fidelity modeling of motion dynamics.

Traditional neural fields methods rely on differentiable volumetric rendering to learn the neural fields from multi-view images. However, this technique often requires densely sampled points along rays to generate high-quality renderings, which makes it unsuitable for real-time applications. For example, TAVA~\cite{li2022tava} and NDF~\cite{zhang2022ndf} require 128 and 64 sample points, respectively, to render a single pixel. To overcome this limitation, we propose a fully differentiable conversion of implicit neural fields to a Neural Explicit Surface (NES) representation that leverages the benefits of both approaches. The conversion enables NES to be trained implicitly and effectively from multi-view videos and be rendered with a new proposed rasterization-based neural renderer that queries texture color only once for the initial ray interaction with the explicit surface, which endows NES with significantly improved inference efficiency and competitive performance with previous neural implicit methods. NES represents the dynamic humans as pose-dependent neural surface deformation fields and their dynamic neural textures both in 2D space, which makes it memory-efficient. We apply the pose-dependent deformation fields to a parametric body model like SMPL~\cite{loper2015smpl} to model geometry dynamics and adopt the dynamic neural textures to render high-fidelity human avatars. By exploiting the above memory efficiency and inference efficiency, NES forwards all pixels in a single MLP pass, making it well-suited for real-time applications.


This paper makes several key contributions, including:
\begin{itemize}[noitemsep]
\item We propose Neural Explicit Surface (NES), a novel representation that models dynamic humans as \textit{pose-dependent neural implicit surface deformation fields} and their \textit{dynamic neural textures} both in 2D space. This results in higher memory and computational efficiency than existing frameworks.
\item We design a fully differentiable conversion between implicit neural fields and the explicit rendering interface of NES. This enables us to train the NES directly from multi-view videos.
\item We propose a rasterization-based neural rendering mechanism that only requires querying the texture color once for the ray intersection with the explicit surface, significantly improving inference and rendering efficiency.
\item We carried out comprehensive experiments and comparisons with existing dynamic human modeling frameworks and validated the memory, fidelity, and scalability of NES.
\end{itemize}


\section{Related Works}

\subsection{Explicit Representation for Dynamic Human Modeling}

Explicit representations, such as voxels~\cite{sitzmann2019deepvoxels, lombardi2019neural}, point clouds~\cite{aliev2020neural, dai2020neural}, and multiplane images (MPI)~\cite{zhou2018stereo, mildenhall2019local}, have been widely used for 3D object representation due to their ease of manipulation, including immediate interactions such as texture editing. However, their memory cost increases significantly with increasing resolution. Parametric models~\cite{anguelov2005scape, loper2015smpl, pavlakos2019expressive} are a prevalent method for fitting parametric body models to skinned human scans. However, they are unsuitable for modeling clothed humans, which have much greater complexity and require expensive 3D scans.

Previous works such as Tex2Shape~\cite{alldieck2019tex2shape} and ~\cite{alldieck2018video} have explored deforming template human vertices to capture more details, including clothing. However, Tex2Shape relies on 3D scans for supervision, while the approach proposed in ~\cite{alldieck2018video} is limited in its ability to represent pose-dependent dynamics. In contrast, our approach involves representing the surface deformation across the entire template surface, and introducing a fully differentiable conversion of implicit neural fields to a Neural Explicit Surface (NES) representation. This facilitates implicit training directly from multi-view videos, enabling the deformation to be dependent on the human pose, allowing us to model pose-dependent dynamics of dynamic humans more accurately.


\subsection{Implicit Representation for Dynamic Human Modeling}

Implicit representations offer potential advantages in memory efficiency and high resolution by representing a scene as continuous functions~\cite{mildenhall2020nerf, yu2021pixelnerf, martin2021nerf, garbin2021fastnerf, liu2020neuralvoxel, yariv2020multiview, wang2021neus, yariv2021volume}. NeRF~\cite{mildenhall2020nerf}, which directly maps a continuous 5D coordinate to the volume density and view-dependent emitted radiance, is a successful example of representing a scene as a neural radiance field. NeRF can represent a continuous scene in arbitrary resolution and effectively learn from multi-view images with a differentiable volumetric renderer.

To represent dynamic humans, recent works have adopted a shared canonical neural implicit representation with a set of deformation fields~\cite{pumarola2021d, park2021nerfies, shao2022doublefield, noguchi2021neural, peng2022animatable, weng2022humannerf, liu2020neural, peng2021animatable}. These methods typically use SMPL~\cite{loper2015smpl} as a prior for skeleton-based motion, and the deformation fields mainly represent non-linear deformations. However, the shared canonical representation can't represent dynamic wrinkles because the wrinkles in the canonical representation are fixed. Alternatively, some works~\cite{zhang2022ndf, xu2022surface} learn to condition the neural implicit representations on the human pose, enabling better representation of pose-dependent dynamics.

Neural implicit representations, however, have limitations. Firstly, they employ volumetric rendering, which defines neural fields in the entire 3D space, despite the 3D object occupying only a small fraction of space. This results in memory redundancy. Secondly, volumetric rendering requires dense sample points along a ray to aggregate as pixel color, resulting in significant computation. In contrast, our Neural Explicit Surface approach represents the 3D object as a 2D deformation map coupled with a 2D texture map, which reduces memory redundancy. The proposed rasterization-based neural rendering only necessitates color querying for one sample point to render a pixel, thus improving computation efficiency significantly.

\begin{figure*}[ht!]
  \centering
  \vspace{-0.3cm}
  \includegraphics[width=1.0\linewidth]{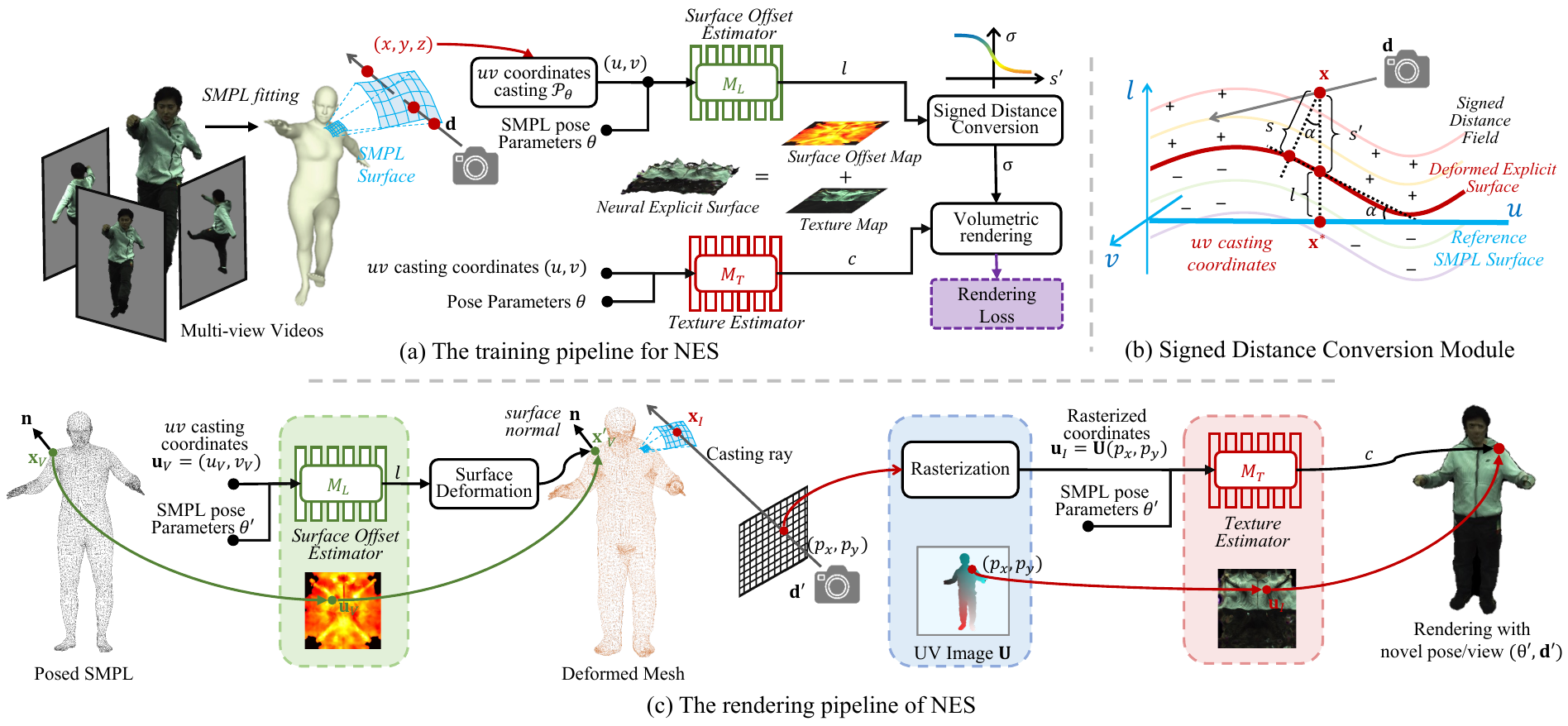}
  \vspace{-0.5cm}
  \caption{The overview of proposed NES. (a) illustrates the NES representation in the center and the corresponding training procedure, as discussed in Section \ref{sec:representation}. (b) shows the proposed fully differentiable conversion between implicit signed distance fields and the deformed explicit surface, which is explained in Section \ref{sec:conversion}. (c) presents the proposed rasterization-based neural rendering algorithm, which is capable of rendering a pixel with just one texture query, as explained in Section \ref{sec:rendering}.
  }
  \vspace{-0.2cm}
  \label{fig:overview}
\end{figure*}


\section{The Proposed Method}
\label{sec: method}

To model dynamic humans from multi-view videos, we adopt the approach proposed in Neural Body~\cite{peng2021neural}, assuming that the cameras are pre-calibrated and synchronized. Off-the-shelf methods are employed to provide the segmentation mask and fitted parametric body model, with SMPL~\cite{loper2015smpl} serving as our parametric body model. In the following sections, we elaborate on our proposed Neural Explicit Surface (NES) representation and its training details in Section~\ref{sec:representation}. We then present the Signed Distance Conversion Module, our proposed fully differentiable conversion between the implicit and explicit representations, in Section~\ref{sec:conversion}. Finally, we present the proposed rendering technique, a rasterization-based neural rendering approach, in Section~\ref{sec:rendering}.


\subsection{Neural Explicit Surface}\label{sec:representation}

To overcome the limitations of current methods for modeling dynamic humans, we propose representing the implicit neural fields as a Neural Explicit Surface (NES). As shown in the center of Fig.~\ref{fig:overview}(a), NES models a dynamic human as a fluttering surface wrapped around an SMPL model, which is then flattened to the NDF space (with $(u,v,l)$ as query coordinate) and decomposed into two learned 2D maps, the Surface Offset map and the Texture map (with $(u,v)$ as query coordinate). The NDF space is borrowed from~\cite{zhang2022ndf}, which projects a 3D scene point 
$(x,y,z)$ onto the NDF-space point $(u,v,l)$ using the SMPL surface as the reference. In this context, $(u,v)$ is the texel coordinate of the closest point on the SMPL surface, and $l$ is the distance offset to the SMPL surface. The Surface Offset Map indicates the distance between the real surface and the reference SMPL surface, while the Texture Map shows the corresponding texture color. 

Decomposing a 3D surface into two 2D maps offers two benefits. First, the 2D representation is more efficient and reduces memory redundancy as the 3D surfaces occupy only a small fraction of the 3D space. Second, by applying the surface offset to the SMPL model, we can immediately obtain the dynamic geometry, avoiding the need for marching cube conversion of implicit neural fields to explicit meshes. This efficient representation of geometry also enables NES to work closely with the graphics pipeline, achieving high inference efficiency, as explained in Section~\ref{sec:rendering}.

To train the NES directly from multi-view videos, we utilize volumetric rendering inspired by NeRF~\cite{mildenhall2020nerf} and integrate it with the fully differentiable conversion presented in Section~\ref{sec:conversion}, as depicted in Fig.~\ref{fig:overview}(a). Firstly, we fit an SMPL to the multi-view video and use it as the reference for UV Coordinate Casting $\mathcal{P_{\theta}}$ (Section~\ref{sec:projection}). The UV Coordinate Casting projects a sample point $(x,y,z)$ to its nearest point on the SMPL surface, and the texel coordinates $(u,v)$ of the closest point are used as the output. The Surface Offset Estimator $M_L$ and Surface Texture Estimator $M_T$ take the texel coordinates $(u,v)$ and the SMPL pose parameters $\theta$ as inputs and output surface offset $l$ and texture color $c$, respectively. We then convert the surface offset $l$ to signed distance $s$ using Signed Distance Conversion Module (Section~\ref{sec:conversion}). The signed distance $s$ is converted into volume density $\sigma$ in accordance with~\cite{wang2021neus}:
\begin{equation}
    \sigma=\left\{
\begin{aligned}
&\frac{1}{\beta}(1-\frac{1}{2} \exp (\frac{s}{\beta})), &\text{if $s < 0$},  \\
& \frac{1}{2 \beta} \exp (- \frac{s}{\beta}), &\text{if $s \geq 0$},
\end{aligned}
\right.
\end{equation}
where $\beta$ is a learnable parameter. Using the converted density, we use standard volumetric rendering with numerical quadrature~\cite{kajiya1984ray} to render pixel color $\hat{\mathbf{C}}(\mathbf{r})$:
\begin{equation}
\hat{\mathbf{C}}(\mathbf{r}) = \sum_{k=1}^{N}{(\prod_{m=1}^{k-1}{e^{-\sigma_m \cdot \delta_m}})\cdot (1-e^{-\sigma_k \cdot \delta_k}) \cdot c_k},
\label{eq:volumerender}
\end{equation}
where $\delta_k=||\mathbf{x}_k-\mathbf{x}_{k-1}||_2$ denotes the quadrature segment, $N$ denotes the number of sample points along a ray, and $c_k$ denotes the predicted color for a sample point. Finally, the rendering loss optimizes the two estimators.

\subsubsection{UV Coordinates Casting $\mathcal{P_{\theta}}$} \label{sec:projection}
To project the sample point $(x,y,z)$ to the nearest point on the SMPL surface and obtain the texel coordinate $(u,v)$, we use the UV Coordinates Casting $\mathcal{P_{\theta}}$. Specifically, given a 3D point $(x,y,z)$, we project it to the nearest point $\mathbf{x^*}=(x^*,y^*,z^*)$ on the SMPL surface using the dispersed projection method presented in Surface-Aligned NeRF~\cite{xu2022surface}. This projection ensures that the projection is injective and finds the most relevant point on the SMPL surface. Then the texel coordinate $\mathbf{u}=(u,v)$ of $\mathbf{x^*}$ is given by:
\begin{align}
  \mathbf{u}=(u,v) \enspace \mathrm{ s.t. } \enspace (x^*,y^*,z^*) = B_{u,v}(\mathcal{V}_{[\mathcal{F}(f)]}),
\end{align}
where $f \in \{1\ldots N_F\}$ is the triangle index, $\mathcal{V}_{[\mathcal{F}(f)]}$ is the three vertices of triangle $\mathcal{F}(f)$, $(u,v): u,v \in [0,1]$ are the texel coordinates on the texture map and $B_{u,v}(\cdot)$ is the barycentric interpolation function.

\subsection{Signed Distance Conversion}\label{sec:conversion}

The Signed Distance Conversion module enables a fully differentiable conversion between the implicit signed distance field and the deformed explicit surface. As illustrated in Figure~\ref{fig:overview}(b), the conversion procedure involves calculating the signed distance of a point to the real surface, which is referred to as the Deformed Explicit Surface in the figure. Specifically, we first project a sample point $\mathbf{x}=(x,y,z)$ onto its closest point $\mathbf{x^*}=(x^*,y^*,z^*)$ on the SMPL surface with texel coordinate $\mathbf{u}=(u,v)$. Then, we input the texel coordinates $\mathbf{u}$ to the Surface Offset Estimator $M_L$, which outputs the surface offset $l$. The surface offset represents the distance between the real surface and the SMPL surface and is positive if the dynamic surface is on the SMPL surface and negative if below the SMPL surface. We calculate the signed distance $s$ of $\mathbf{x}$ using the surface offset $l$ as follows:
\begin{align}\label{eq:conversion}
s = ((\mathbf{x} - \mathbf{x^*}) - l\cdot\mathbf{n})\cdot \cos(\alpha),
\end{align}
where $\mathbf{n}$ is the SMPL surface normal vector at $\mathbf{x^*}$, and $\alpha$ is the angle between the real surface and the SMPL surface. We utilize the angle $\alpha$ to refine the distance $s'$ perpendicular to the SMPL surface as the distance $s$ perpendicular to the real surface, which results in a more accurate computation. To compute $\alpha$, we derive $\tan(\alpha)$ as the length of the gradient of the surface offset $l$, which is computed using the auto gradient function of Pytorch.

\subsection{Rasterization-based Neural Renderer}\label{sec:rendering}

The proposed rasterization-based neural rendering technique efficiently renders the dynamic human under specific SMPL pose parameters and camera settings. As shown in Fig.~\ref{fig:overview}(c), to render the performer under specific SMPL pose parameters $\theta'$ and camera $\bf{d'}$, we obtain the dynamic geometry by deforming the SMPL vertices with the Surface Deformation Function:
\begin{equation}\label{eq:deform}
\mathbf{x'}_V=\mathbf{x}_V + l \cdot \mathbf{n},
\end{equation}
where $\mathbf{x}_V$ is a vertex of SMPL, $\mathbf{n}$ is the normal for $\mathbf{x}_V$, $l$ is the surface offset predicted by the Surface Offset Estimator $M_L$, and $\mathbf{x'}_V$ is the deformed vertex. We then combine the deformed vertices and the triangle relation defined by SMPL to create the Deformed Mesh. This mesh, along with the camera parameters $\bf{d'}$, is input to the rasterization process to obtain mesh property for each pixel. Note the Surface Offset Estimator $M_L$ is defined over the whole SMPL surface. However, for simplicity, we only use $M_L$ to deform the SMPL vertices and find it works well in experiments.

We adopt the rasterization implemented with pytorch3d~\cite{liu2019soft,ravi2020accelerating} and set the mesh property for each pixel as the texel coordinates for the first intersection of the pixel ray with the Deformed Mesh. The output is the UV image $\bf{U}$, which records the texel coordinate $\mathbf{u}_I=\mathbf{U}(p_x,p_y)$ for each image-space coordinate $(p_x,p_y)$. Finally, we use the Surface Texture Estimator $M_T$ to render the UV image into a color image by querying pixel color for each image-space coordinates $(p_x,p_y)$ with its texel coordinates $\mathbf{u}_I$ as input to $M_T$. The estimators $M_L$ and $M_T$ are both implemented with MLPs, and the SMPL pose parameters $\theta'$ are used to condition the estimators and represent pose-dependent shape and texture dynamics.

Our proposed rasterization-based neural rendering approach has three key advantages:

1) our approach queries the texture color only once for rendering a pixel, which is a significant reduction from tens of sample points used in previous methods~\cite{peng2021neural,zhang2022ndf}. This decrease in sample points allows feeding all pixels into the estimators simultaneously, thus improving rendering efficiency.

2) as the Surface Offset and Texture are conditioned on SMPL pose parameters, NES can represent pose-dependent dynamics, thus enabling more accurate rendering of human motion.

3) as the Surface Offset and Texture are learned in 2D space with texel coordinates as input, our approach is much more compact compared to previous works that learn in 3D space, thus reducing memory redundancy.

\section{Experiment}\label{sec: experiment}

\subsection{Implementation Details}

To achieve accurate and efficient mesh deformation, we implement the Surface Offset Estimator $M_L$ and Texture Estimator $M_T$ using Multi-Layer Perceptron (MLP). For training, we adopt a single-stage sampling strategy with 128 samples and an additional 16 points around the first ray intersection with the deformed mesh to ensure accurate surface estimation. MSE loss of pixel colors is used to optimize the two estimators.

Notably, our method outperforms ARAH~\cite{wang2022arah}, which also adopts a similar sampling strategy. ARAH calculates the ray intersection with the implicit Signed Distance Function (SDF) using the Joint Root-Finding algorithm. This computation-heavy process takes around 1 second at inference, making it unsuitable for real-time applications. In contrast, our NES uses a rasterization-based rendering approach that is faster and more efficient.


\subsection{Experimental Settings}
\textbf{Dataset and Metrics.} To evaluate the effectiveness of our proposed method, we conduct experiments on the ZJU-MoCap dataset~\cite{peng2021neural}, which records multi-view videos with 23 synchronous cameras and collects the shape and pose parameters of SMPL. Following the same protocol as~\cite{peng2021neural}, we randomly choose 4 cameras for training and use the remaining cameras for testing. Our goal is to evaluate the novel view rendering and novel pose animation tasks using two commonly used metrics: PSNR as an image-based metric and LPIPS as a perceptual metric. LPIPS is considered a more representative measure of visual quality, as it takes into account the perceptual similarity between the generated images and the GT images.

\textbf{Compared Methods.} We benchmark our NES against three SOTA methods: AniSDF~\cite{peng2022animatable}, NDF~\cite{zhang2022ndf} and SANeRF~\cite{xu2022surface}. AniSDF represents the 3D model with canonical signed distance fields and learns frame-specific deformation fields to aggregate information across frames to the shared canonical model. However, it has limited pose-dependent dynamics due to the fixed texture of the canonical model. AniSDF is an extended version of AniNeRF~\cite{peng2021animatable} that replaces the neural radiance fields in AniNeRF with signed distance fields, leading to better performance. In contrast, NDF and SANeRF represent the scene with neural radiance fields that are conditioned on the human pose, which shows better pose-dependent dynamics. 

All the compared methods and other SOTA methods~\cite{li2022tava,geng2023learning,wang2022arah} model the human body as implicit fields, which requires dense sample points during inference. In contrast, our proposed NES method adopts a rasterization-based neural rendering approach, which only requires a single sample point along each pixel ray. This results in significantly faster and more efficient inference, making our method more suitable for real-time applications.


\subsection{Rendering Efficiency}

\begin{table}[h!]
\begin{center}
\vspace{-0.5cm}
    \caption{Rendering efficiency.}
    \vspace{-0.4cm}
    \label{tab:efficiency}
    \resizebox{0.95\linewidth}{!}{
    \begin{tabular}{c|c|c|c}
                    & memory(Gb) &  LPIPS $\times10^3$ $\downarrow$ &  FPS $\uparrow$\\
    \Xhline{1.2pt}
    SANeRF~\cite{xu2022surface}	        &8.3	    &30.65	    & 0.36\\
    ARAH~\cite{wang2022arah}	        &22.1	    &\textbf{25.02}	& 0.09\\
    NeuralBody~\cite{peng2021neural}	    &5.2	    &34.87	& 3.59\\
    AniSDF~\cite{peng2022animatable}          & 11.4     & 42.15     & 0.93  \\
    NDF~\cite{zhang2022ndf}             & 5.5      & 37.30     & 2.48  \\
    \hline
    NES\_VR         & 19.3     & 28.50     & 0.12  \\
    NES\_VR\_s	&14.6	    &30.02	& 0.13\\
    NES\_VR\_m	&13.1	    &27.78	& 0.13\\
    NES\_VR\_UVL    & 34.6     & 36.37     & 0.13  \\
NES\_VR\_UVL\_s	&21.5	    &35.80	& 0.13\\
NES\_VR\_UVL\_m	&16.0	    &39.65	& 0.13\\
    \textbf{\color{red}{NES\_RR}}         & 3.2      & \textbf{26.96} & 15.89 \\
    NES\_RR\_s	&3.2	    &27.39	& 16.22\\
    NES\_RR\_m	&3.2	    &27.96	& 16.82\\    
    \end{tabular}
    }
    \vspace{-0.4cm}
\end{center}
\end{table}

The rendering efficiency of NES is one of its most outstanding advantages in representing dynamic humans in real-time applications. To evaluate the efficiency of our proposed method, we compare it with several previous methods, including NDF~\cite{zhang2022ndf}, AniSDF~\cite{peng2022animatable}, SANeRF~\cite{xu2022surface}, ARAH~\cite{wang2022arah}, NeuralBody~\cite{peng2021neural}, and several variants of NES, where `s', `m' mean different MLP layers, `VR' means Volume Rendering, `RR' means Rasterization-based Neural Rendering, and `UVL' means replacing the 2D input coordinate with a 3D input coordinate to be more similar to previous methods defined in 3D space. Section~\ref{sec:ablation} depicts the variants in more detail.

To evaluate the rendering efficiency, we render an image of $512 \times 512$ resolution 100 times and take the mean rendering time. We take the LPIPS on subject 387 of ZJU-MoCap as the rendering performance metric. We run the comparison on a machine with 64 AMD EPYC 7302 16-Core processors and 2 A100 GPUs. The results, shown in Table~\ref{tab:efficiency} and Figure~\ref{fig:efficiency}, demonstrate that our proposed NES with Rasterization-based Rendering (NES\_RR entry) produces the best trade-off between performance and efficiency as well as the memory cost. While ARAH produces slightly better LPIPS than our NES\_RR, its memory cost is 7 times of ours, and its rendering speed is over 160 times slower than our NES\_RR. Notably, with the same network design and excluding the implementation tricks, our proposed Rasterization-based Neural Rendering improves the FPS from 0.12 (NES\_VR entry with Volumetric Rendering) to 15.89 (NES\_RR entry with Rasterization-based Rendering), representing over 130 times speedup. These results demonstrate the significant advantage of our proposed NES method in efficiently representing dynamic humans in real-time applications.


\subsection{Quantitative Results}
\label{quantity}

\begin{table*}[ht]\small
\begin{center}
    \caption{Quantitative results of novel view rendering and novel pose animation on the ZJU-MoCap dataset in terms of PSNR (higher is better) and LPIPS (lower is better). The best result is \textbf{highlighted}.}
    \vspace{-0.3cm}
    \label{tab:zjumocal_nvnp}
    \resizebox{0.95\textwidth}{!}{
    \begin{tabular}{c|c|c|c|c|c|c|c|c|c|c|c|c|c|c|c|c}
        & \multicolumn{8}{c|}{Novel View Rendering} &  \multicolumn{8}{c}{Novel Pose Animation}\\
        & \multicolumn{4}{c|}{PSNR $\uparrow$} &  \multicolumn{4}{c|}{LPIPS $\times 10^3 \downarrow$} &  \multicolumn{4}{c|}{PSNR $\uparrow$} &  \multicolumn{4}{c}{LPIPS $\times 10^3 \downarrow$}\\
    \Xhline{1.2pt}
        & AniSDF &  NDF  & SANeRF & OURS           & AniSDF &  NDF  & SANeRF  & OURS           & AniSDF &  NDF  & SANeRF & OURS             & AniSDF &  NDF  & SANeRF  & OURS   \\                     
    \hline
    377 & \textbf{27.44}  & 26.84 & 27.13  & 27.01          & 21.41  & 23.23 & \textbf{19.01}   & 19.62          & \textbf{25.14}  & 24.59 &  24.87 & 25.13            & 23.51  & 26.32 &  \textbf{22.22}  & 22.74  \\
    386 & 28.29  & 28.21 & \textbf{28.58}  & 28.07          & 27.66  & 22.05 & \textbf{18.10}   & 22.27          & 26.77  & 26.32 &  26.43 & \textbf{27.20}            & 32.19  & 28.93 &  \textbf{24.89}  & 27.48  \\
    387 & 25.71  & 24.52 & 25.93  & \textbf{26.06}          & 42.15  & 37.31 & 30.65   & \textbf{26.96}          & 23.33  & 22.40 &  22.84 & \textbf{23.60}            & 43.73  & 40.73 &  36.85  & \textbf{29.75}  \\
    392 & 28.81  & 28.39 & \textbf{29.03}  & 28.09          & 32.95  & 29.87 & \textbf{22.56}   & 23.90          & 25.66  & 25.61 &  25.49 & \textbf{26.26}            & 39.96  & 39.05 &  33.03  & \textbf{31.06}  \\
    393 & 26.74  & 26.73 & \textbf{27.18}  & 26.78          & 33.11  & 30.98 & \textbf{25.56}   & 25.63          & 23.47  & 24.02 &  24.02 & \textbf{24.50}            & 44.64  & 42.66 &  39.10  & \textbf{38.60}  \\
    394 & 28.33  & 27.97 & \textbf{28.68}  & 27.98          & 33.10  & 29.94 & \textbf{23.44}   & 26.86          & 24.44  & 24.29 &  24.64 & \textbf{25.44}            & 41.67  & 40.70 &  34.70  & \textbf{33.43}  \\
    \hline
    Avg.& 27.55  & 27.11 & \textbf{27.75}  & 27.33          & 31.73  & 28.89 & \textbf{23.22}   & 24.20          & 24.80  & 24.53 &  24.71 & \textbf{25.35}            & 37.61  & 36.39 &  31.79  & \textbf{30.51}  \\
    \end{tabular}
    }
    \vspace{-0.3cm}
\end{center}
\end{table*}

Table~\ref{tab:zjumocal_nvnp} presents the quantitative comparison of novel view rendering and novel pose animation tasks for the proposed NES and compared methods, including AniSDF~\cite{peng2022animatable}, SANeRF~\cite{xu2022surface} and NDF~\cite{zhang2022ndf}. With different implementation details, the proposed NES slightly lags behind previous works on the novel view rendering task but outperforms them on the novel pose animation task by 0.55 of PSNR and 1.2 of LPIPS, demonstrating its superior performance in producing visually high-quality results. Moreover, NES achieves these results with significantly lower memory cost and faster rendering speed, as shown in Table~\ref{tab:efficiency}. 

The overtake of NES demonstrates that NES has a smaller performance gap between the training pose and the unseen pose when compared to previous methods. This better generalization is a result of the compact learnable space of NES, which learns the surface offset and texture in 2D space with the texel coordinate as input. In contrast, previous methods learn in 3D space with 3D coordinates as input, resulting in a much more complex learnable space. Overall, these findings demonstrate that the proposed NES method is highly effective in representing dynamic humans for both novel view rendering and novel pose animation tasks.

\subsection{Qualitative Results}

\begin{figure*}[h!]
  \centering
  \includegraphics[width=0.95\linewidth]{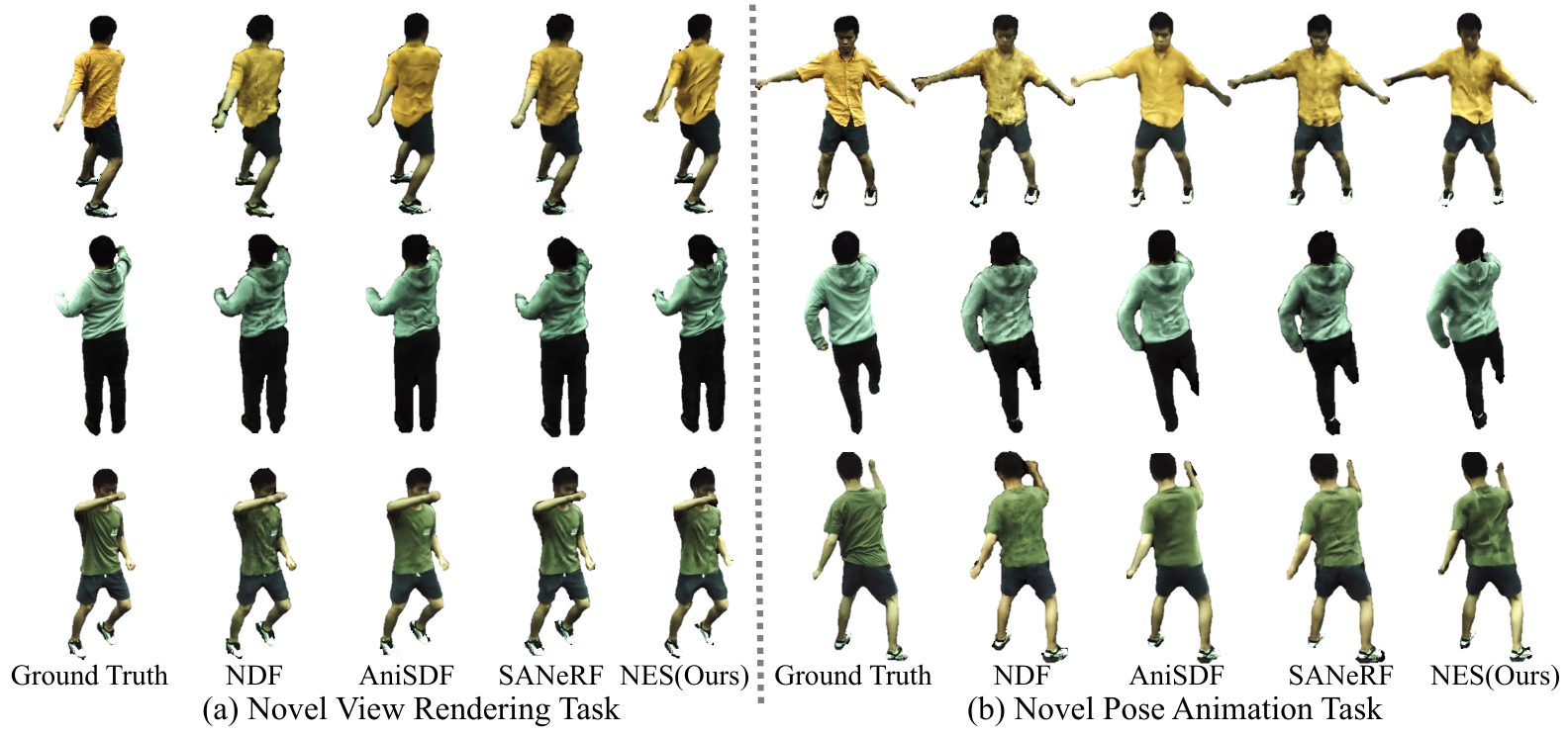}
\vspace{-0.4cm}
\caption{Qualitative comparison with SOTA methods in (a) novel view rendering task, and (b) novel pose animation task on the ZJU-MoCap dataset.}
\label{fig:comparison_nvnp}
\end{figure*}

Figure~\ref{fig:comparison_nvnp} presents a qualitative comparison of the proposed NES with compared methods, NDF~\cite{zhang2022ndf}, SANeRF~\cite{xu2022surface} and AniSDF~\cite{peng2022animatable}, for both novel view rendering and novel pose animation tasks. The results demonstrate that NES outperforms the compared methods in recovering clean and evident pose-dependent dynamics, particularly in the representation of wrinkles, folds, and clothing creases. For instance, NES successfully recovers the wrinkles on the clothes of the first subject in the first row of Fig.~\ref{fig:comparison_nvnp}, while the wrinkles produced by NDF are blurry and dirty, as shown in the first row and second column of the same figure. In comparison, AniSDF produces over-smoothed renderings or unreasonable wrinkles, as shown in the third column of the first and second rows in Fig.~\ref{fig:comparison_nvnp}.

Moreover, the proposed NES demonstrates better generalization to unseen poses, which is due to its more compact learnable space. Specifically, NES learns surface offset and texture in 2D space, with texel coordinates as input. In contrast, compared methods are defined in 3D space, resulting in a much larger learnable space. As a result, NES successfully produces wrinkles of the first performer on the first row (the rightmost column). In comparison, though SANeRF produces reasonable wrinkles for the same performer under training poses (the fourth column), it degenerates greatly when applied to novel poses (the ninth column). These findings demonstrate the benefits of the proposed NES method in efficiently representing dynamic humans with high-quality results.


\subsection{Ablations}\label{sec:ablation}

\begin{figure*}[h!]
  \centering
  \includegraphics[width=0.95\textwidth]{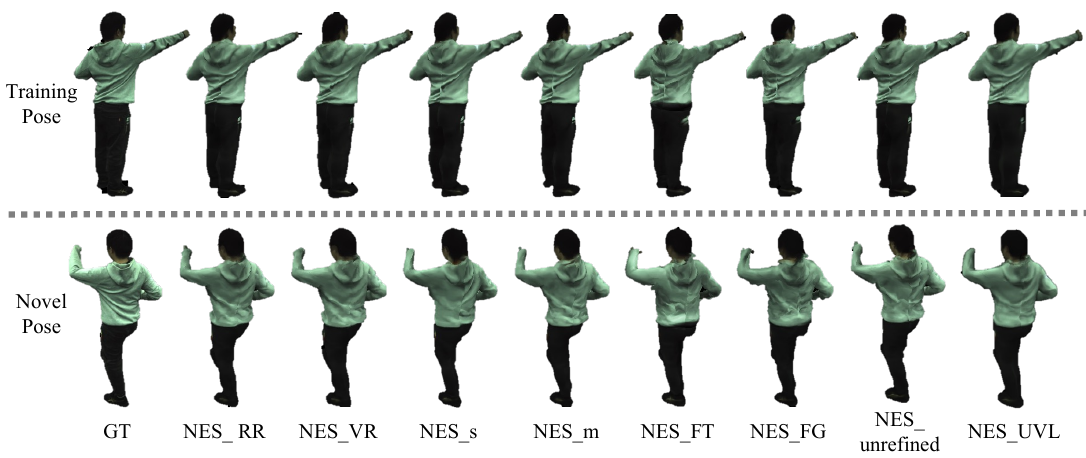}
  \vspace{-0.4cm}
\caption{Ablation studies.}
\label{fig:ablation}
\end{figure*}

\begin{table*}[ht]
  \begin{center}
    \caption{Quantitative result of ablation study.}
    \label{tab:ablation}
    \resizebox{0.99\textwidth}{!}{
    \begin{tabular}{c|c|p{1.3cm}<{\centering}|p{1.3cm}<{\centering}|p{1.3cm}<{\centering}|p{1.3cm}<{\centering}|p{1.3cm}<{\centering}|p{1.3cm}<{\centering}|p{1.3cm}<{\centering}|p{1.3cm}<{\centering}}
    \multicolumn{2}{c|}{Ablation Item}                      & NES     &  NES      & \makecell[c]{NES\_s}& \makecell[c]{NES\_m}  &  \makecell[c]{NES\_FT} & \makecell[c]{NES\_FG} & \makecell[c]{NES\_\\unrefined}  & \makecell[c]{NES\_UVL}  \\
      \Xhline{1.2pt}
    \multicolumn{2}{c|}{Rendering Method}                   & RR      &  VR       & RR         &  RR          &  RR & RR  &  RR  &  VR  \\
      \cline{1-10}
    \multirow{3}*{\makecell[c]{Seen Pose}}          & PSNR $\uparrow$  & 26.061  &  26.347   & 25.987     &  25.933      &  25.242 & 25.004  &  26.010  & \textbf{26.765}  \\
                                                    & SSIM $\uparrow$  & \textbf{0.930}   &  0.928    & 0.929      &  0.928       &  0.914 & 0.915  &  0.929  &  0.929  \\
                                                    & LPIPS $\downarrow$ & \underline{\textbf{26.961}}  &  28.507   & 27.399     &  27.966      &  33.564 & 32.397  &  27.492  &  36.376  \\
    \cline{1-10}  
    \multirow{3}*{\makecell[c]{Unseen Pose}}        & PSNR $\uparrow$  & 23.602  &  23.957   & 23.614     &  23.603      &  23.162 & 23.316  &  23.591  &  \textbf{24.190}  \\
                                                    & SSIM $\uparrow$  & 0.900   &  0.899    & 0.901      &  0.901       &  0.896 & 0.895  &  0.900  &  \textbf{0.902}  \\
                                                    & LPIPS $\downarrow$ & \underline{\textbf{29.751}}  &  31.690   & 29.952     &  30.163      &  34.179 & 32.918  &  30.138  &  39.684  \\
    \Xhline{1pt}
    \end{tabular}
    }
  \end{center}
\end{table*}

To validate the effectiveness of the proposed NES, we conduct extensive ablation studies, including the following entries:
\begin{itemize}[noitemsep]
\item NES\_RR adopts Rasterization-based Neural Rendering (RR), which only queries colors once for each pixel.
\item NES\_VR adopts Volumetric Rendering (VR), which follows the training stage and requires 144 sample points for a pixel.
\item NES\_s reduces MLP layers in half. Specifically, the original NES has 12-layer MLP, and NES\_s has 6 layers.
\item NES\_m reduces the number of MLP layers to 3.
\item NES\_FT has Fixed Textures (FT) by not depending the Surface Texture Estimator $M_T$ on the human pose.
\item NES\_FG has Fixed Geometry (FG) by not depending the Surface Offset Estimator $M_L$ on the human pose.
\item NES\_unrefined directly adopts the $s'$ in Figure~\ref{fig:overview}(b) as signed distance without the refinement of $cos(\alpha)$.
\item NES\_UVL adopts 3D coordinates $(u,v,l)$ as query input of neural fields, which is similar to NDF~\cite{zhang2022ndf} and represents traditional neural fields methods defined in 3D space.
\end{itemize}

\subsubsection{Effect of Rasterization-based Neural Rendering}\label{sec:ablation_RR}
We adopt Volumetric Rendering (VR) for training and Rasterization-based Neural Rendering (RR) for inference. Table~\ref{tab:efficiency} and Figure~\ref{fig:efficiency} have shown a significant boost in rendering efficiency brought by Rasterization-based Neural Rendering. The second and third column in Table~\ref{tab:ablation} and Figure~\ref{fig:ablation} shows a more detailed comparison, indicating that Rasterization-based Neural Rendering (RR) increases the LPIPS metric of NES(RR) but decreases the PSNR metric. We attribute this to the different geometry representations. On the one hand, the geometry of Rasterization-based Neural Rendering (RR) is the deformed SMPL surface, which is relatively coarse and may cause pixel misalignment with the ground truth image. Thus the pixel-wise metric PSNR is reduced.
On the other hand, Rasterization-based Neural Rendering (RR) only queries colors for one sample point, thus alleviating the stratified effect caused by sparse sample points, as shown in the third column and first row of Figure~\ref{fig:ablation}, where the wrinkles on the back have a stratified effect. In this way, the perceptual metric LPIPS is improved. We focus on the image synthesis task, and the LPIPS metric is more representative of visual quality. 

\subsubsection{Memory Efficiency}
To validate the memory efficiency, we ablate NES with fewer layers of MLP. NES has 12 layers, NES\_s has 6 layers, and NES\_m has 3 layers. The fourth and fifth columns in Table~\ref{tab:ablation} and Figure~\ref{fig:ablation} show the result. With fewer layers, the LPIPS metric slowly degrades but is still competitive. The fourth and fifth columns in Figure~\ref{fig:ablation} show that the performance of image synthesis is stable. This study indicates that NES is memory efficient and can be further compressed to improve inference efficiency.

\subsubsection{Pose-dependent Dynamics}
NES\_FG has Fixed Geometry (FG), and NES\_FT has Fixed Textures (FT). This study is to show that NES can recover pose-dependent dynamics. The sixth and seventh columns in Table~\ref{tab:ablation} and Figure~\ref{fig:ablation} indicate that all three metrics decrease drastically, and the synthesized image in Figure~\ref{fig:ablation} shows limited and blurred dynamics, like the second row, where the wrinkles on the back disappear compared with the GT image.

\subsubsection{Effect of Refinement Equation}
The unrefined version of NES is shown as NES\_unrefined entry in Table~\ref{tab:ablation} and Figure~\ref{fig:ablation}, indicating that the refinement improves the LPIPS by 0.5 for training poses and 0.4 for unseen poses. The visual comparison shows that the refined version of NES (the second column NES\_RR in Figure~\ref{fig:ablation}) produces better pose-dependent dynamics, e.g., the wrinkles of the second row and the second column are cleaner. NES\_RR also shows better generalization to novel poses, e.g., the last row of the second column in Figure~\ref{fig:ablation} produces reasonable wrinkles, but the eighth column does not.

\subsubsection{Effect of Compression to 2D Space}
We represent NES in 2D space, and prevalent neural field methods are defined in 3D space. To study the difference, we replace the input of NES from 2D coordinate $(u,v)$ to 3D coordinate $(u,v,l)$ and keep other designs unchanged, which is denoted as NES\_UVL in Table~\ref{tab:ablation} and Figure~\ref{fig:ablation}. Note that NES\_UVL cannot be rendered with Rasterization-based Neural Rendering and is over 100 times slower than NES\_RR. The last column of Table~\ref{tab:ablation} and Figure~\ref{fig:ablation} shows NES\_UVL produces higher PSNR and worse LPIPS. The reason is similar to Section~\ref{sec:ablation_RR}, i.e., the coarse geometry of NES\_RR causes pixel misalignment. Thus, the PSNR is reduced. And the stratified effect is alleviated; therefore, the LPIPS is improved. For example, the first row of the last column in Figure~\ref{fig:ablation} contains the stratified effect on the back wrinkles.

\subsection{Texture Editing}
\begin{figure}[h!]
  \centering
  \includegraphics[width=0.8\linewidth]{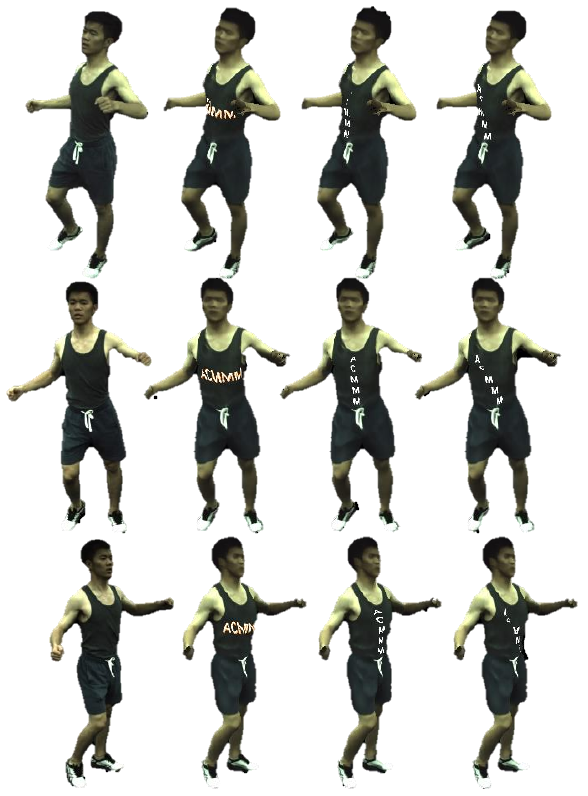}
\caption{Texture Editing. The leftmost column is the ground truth without texture editing.}
\label{fig:texture}
\end{figure}
With the explicit representation of the texture map with 2D $(u,v)$ as query coordinates, we can efficiently conduct texture editing to the learned NES, as shown in Figure~\ref{fig:texture}. Specifically, in the Rasterization-based Neural Rendering, after getting the texel coordinates for each pixel, we query the color with the texel coordinates as input from the Texture Estimator $M_T$, as well as from a drawn mask with a target pattern on the texture map, and fuse the two queried colors.

\section{Limitations}

Despite the positive findings of NES in efficiently representing dynamic humans with high-quality results, there are still some limitations that need to be addressed in future research. For example, the current design of only deforming the SMPL with a single layer may fail under special circumstances, such as loose skirts where the clothing is not appressed to the skin. Future research could investigate more advanced parametric clothed human models that capture such phenomena to improve the realism of NES.

Another limitation of the proposed NES is its limited performance in recovering hand details, which is due to the lack of hand fitting in the SMPL model. Future research could explore the cooperation of NES with stronger parametric body models, such as SMPL-X~\cite{pavlakos2019expressive}, to capture more details on hands and improve the overall quality of the produced animations.

Despite these limitations, the proposed NES is a significant improvement over previous methods and has the potential to enable various real-time applications, including virtual try-on, telepresence, and gaming. Future research could explore extending NES's capabilities to other types of objects, such as animals or inanimate objects, further expanding its potential for practical use cases.


\section{Conclusion}
\label{sec: conclusion}

The incorporation of NES with the graphics pipeline allows for querying only one sample point per ray, greatly reducing the computational cost of rendering in real-time applications. NES also improves the generalization of novel pose animations by reducing the performance gap between training and unseen poses. These capabilities represent important contributions to the field of computer graphics, which will enable more realistic and efficient real-time dynamic human avatars. Overall, NES represents a significant step forward in the field of computer graphics, and we look forward to the innovative applications that will likely result from this breakthrough research.


\section*{Acknowledgments}
The research was supported by the Theme-based Research Scheme, Research Grants Council of Hong Kong (T45-205/21-N).

\bibliographystyle{ACM-Reference-Format}
\balance
\bibliography{egbib}

\end{document}